\documentclass[runningheads]{llncs}
\usepackage{graphicx}
\usepackage{comment}
\usepackage{amsmath,amssymb} 
\usepackage{color}

\usepackage{subfigure}
\usepackage{makecell}
\usepackage{slashbox}
\usepackage{algorithm}
\usepackage{algorithmic}

\begin{document}
\pagestyle{headings}
\mainmatter

\title{Progressively Unfreezing Perceptual GAN} 

\titlerunning{PUPGAN}
%
\author{Jinxuan Sun\inst{1} \and
Yang Chen\inst{1} \and
Junyu Dong\inst{1} \and
Guoqiang Zhong\inst{1}}
\authorrunning{J. Sun et al.}
%
\institute{Ocean University of China\\ 
\email{gqzhong@ouc.edu.cn}}
\maketitle\vspace{-0.5cm}

\begin{abstract}
Generative adversarial networks (GANs) are widely used in image generation tasks, yet the generated images are usually lack of texture details. In this paper, we propose a general framework, called Progressively Unfreezing Perceptual GAN (PUPGAN), which can generate images with fine texture details. Particularly, we propose an adaptive perceptual discriminator with a pre-trained perceptual feature extractor, which can efficiently measure the discrepancy between multi-level features of the generated and real images. In addition, we propose a progressively unfreezing scheme for the adaptive perceptual discriminator, which ensures a smooth transfer process from a large scale classification task to a specified image generation task. The qualitative and quantitative experiments with comparison to the classical baselines on three image generation tasks, i.e. single image super-resolution, paired image-to-image translation and unpaired image-to-image translation demonstrate the superiority of PUPGAN over the compared approaches.
\keywords{Generative adversarial networks, Fine texture details, Image generation, Progressively unfreezing}
\end{abstract}

\section{Introduction}\label{sec:intro}

In recent years, generative adversarial networks (GANs) \cite{DBLP:journals/corr/GoodfellowPMXWOCB14} have been widely applied to numerous types of image generation tasks, such as single image super-resolution \cite{DBLP:conf/cvpr/LedigTHCCAATTWS17,DBLP:conf/eccv/WangYWGLDQL18}, image-to-image translation \cite{DBLP:conf/iccv/ZhuPIE17,DBLP:conf/cvpr/IsolaZZE17}, image inpainting \cite{DBLP:conf/cvpr/Yu0YSLH18,DBLP:conf/cvpr/Yeh0LSHD17} and image deblurring \cite{DBLP:conf/cvpr/KupynBMMM18,DBLP:conf/cvpr/LuCC19}. In a nutshell, GANs are a framework to produce a generated distribution to match a given target distribution. The architecture of GANs consists of a generator that produces the generated distribution and a discriminator that evaluates the discrepancy between the generated and real data distributions. The generator and the discriminator are trained alternatively with the adversarial loss. To improve the quality of the generated images, the perceptual loss \cite{DBLP:conf/eccv/JohnsonAF16} is usually used to measure the discrepancy of the high-level features on image generation tasks \cite{DBLP:conf/cvpr/LedigTHCCAATTWS17,DBLP:conf/eccv/WangYWGLDQL18,DBLP:journals/corr/ZhangSP17,DBLP:conf/cvpr/KupynBMMM18}. Nevertheless, the perceptual loss is measured by an external pre-trained convolutional neural network. It generally employs the VGGNet-16/19 \cite{DBLP:journals/corr/SimonyanZ14a} pre-trained on the ImageNet dataset \cite{DBLP:journals/ijcv/RussakovskyDSKS15}. In this case, the perceptual loss mainly focuses on the high-level features that contribute to the specific classification task, and therefore, may perform inferiorly on the other tasks. Moreover, utilizing an external network as the feature extractor ignores the fact that the discriminator in GANs can learn the representations of the images and measure the discrepancy between them.

In addition, there is a persisting challenge in the training of GANs \cite{DBLP:conf/nips/SalimansGZCRCC16,DBLP:conf/iclr/ArjovskyB17}. When the generated distribution and the real data distribution are perfectly distinguished by the discriminator, the training of the generator comes will be stopped, as the gradient produced by the discriminator is $0$. A typical cause of this issue is that the discriminator rapidly overpowers the generator. To address this problem, Sajjadi et al. \cite{DBLP:conf/icml/SajjadiPMS18} propose a module to degrade the real data before feeding them to the discriminator, which balances the training of the generator and discriminator. However, this method greatly slows down the learning speed of the discriminator and the whole network.

In this paper, we propose a general framework, named Progressively Unfreezing Perceptual GAN (PUPGAN), which can generate images with fine texture details. Particularly, we propose an adaptive perceptual discriminator architecture, which utilizes a pre-trained dense block~\cite{DBLP:conf/cvpr/HuangLMW17} as a perceptual feature extractor, and the capability encoded in the percetual feature extractor can be transferred to the current task. The last layer in the feature extractor obtains multi-level features from all preceding layers. Following the perceptual feature extractor, the discriminative learning layers are used to measure the discrepancy between the multi-level features of the generated and real images. In addition, we propose a progressively unfreezing scheme to stabilize the training of PUPGAN. Specifically, we unfreeze the parameters in the perceptual feature extractor layer-by-layer during the training process.

\begin{figure}[t]
   \centering
   \includegraphics[width=\textwidth]{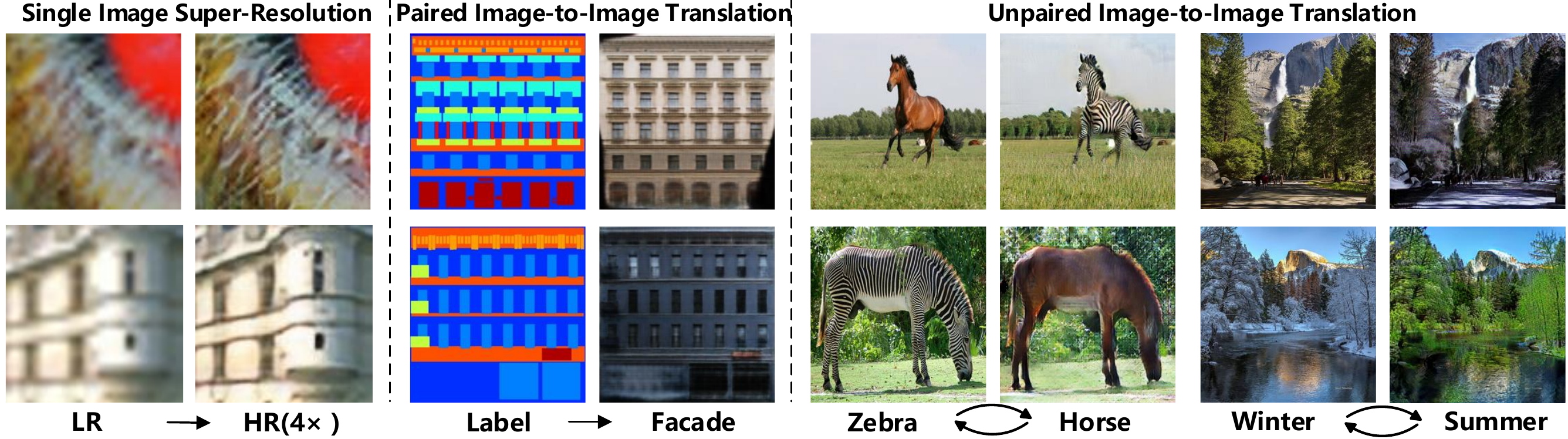}
   \caption{The images generated by PUPGAN: ($left$) for single image super-resolution; ($center$) for paired image-to-image translation; ($right$) for unpaired image-to-image translation.}
   \label{fig:first}
\end{figure}
In summary, we make the following contributions in this work:\vspace{-0.3cm}
\begin{itemize}
\item We propose a general framework, called PUPGAN, for image generation with fine texture details. It can be applied to various scenarios to improve the texture details of the images generated by GANs-based models.
\item We propose an adaptive perceptual discriminator to measure the discrepancy between multi-level features of the generated and real images. In contrast to traditional perceptual loss based on an external network, we fully exploit the capability of the discriminator for multi-level feature extraction.
\item We propose a progressively unfreezing scheme for PUPGAN, to smoothly transfer external knowledge learned on large scale applications to the current task. 
\item We have demonstrated the effectiveness of PUPGAN on several tasks, as shown in Fig.~\ref{fig:first}. In particular, we have applied PUPGAN to unpaired image-to-image translation and obtained promising results. To our best, this is the first work to efficiently exploit the external knowledge and adapt it to the GANs-based unpaired image-to-image translation.
\end{itemize}
\section{Related Work}\label{sec:rw}
Since Goodfellow et al. \cite{DBLP:journals/corr/GoodfellowPMXWOCB14} introduced GANs, they have received more and more attention from the deep learning community. Radford et al. \cite{DBLP:journals/corr/RadfordMC15} modified the architecture of GANs with convolutional layers \cite{DBLP:conf/icml/IoffeS15} to improve their performance and stabilize their training. Following this methodology, Salimans et al. \cite{DBLP:conf/nips/SalimansGZCRCC16} proposed several techniques to encourage the convergence of GANs, such as feature matching and minibatch discrimination.

To further improve the learning capability of GANs, the researchers proposed to train GANs with additional loss functions, such as pixel-wise loss and perceptual loss \cite{DBLP:conf/eccv/JohnsonAF16}. The reason behind this is that additional measurement can help to better evaluate the discrepancy between the generated image distribution and the real one. Specifically, the perceptual loss encourages the generated images to have similar high-level features with the real image. Though integrating perceptual loss to GANs has produced impressive results, it depends on the external convolutional neural network (e.g., VGGNet-19 \cite{DBLP:journals/corr/SimonyanZ14a}) trained on a specific classification task. As a result, the perceptual loss mainly focuses on high-level features relevant to the original dataset, e.g., ImageNet~\cite{DBLP:journals/ijcv/RussakovskyDSKS15}. Recently, Li et al. \cite{DBLP:conf/cvpr/LiLWXFY17} proposed a perceptual GAN for small object detection. The discriminator of this perceptual GAN consists of a perception branch. In addition, Wang et al. \cite{DBLP:journals/tip/WangXWT18} proposed a perceptual adversarial loss to train an image-to-image transformation network. The VGGNet was adopted as an internal perception function in the discriminator. Moreover, Sungatullina et al. \cite{DBLP:conf/eccv/SungatullinaZUL18} proposed a perceputal discriminator, which embeded a pre-trained feature extractor (VGGNet) inside the discriminator. However, the parameters in the pre-trained feature extractor were fixed during training. These approaches have improved the quality of the generated image to some extent. Nevertheless, the high-level features contribute to the specific classsication task and may perform inferiorly on the other tasks. In this work, we embed a well-trained dense block into the discriminator of the proposed PUPGAN. With the fine-tuning of the adaptive perceptual discriminator, PUPGAN can sufficiently leverage the external knowledge and adapt to the current task.

In order to overcome the unstable training problem of GANs due to  the discriminator overpowering the generator, some efforts have been made. Sajjadi et al. \cite{DBLP:conf/icml/SajjadiPMS18} introduced a convolutional model to balance the generator and discriminator by gradually revealing the details of the real data. This method encourages a smooth training procedure. However, it slows down the training speed of the entire network. Karras et al. \cite{DBLP:conf/iclr/KarrasALL18} proposed to grow the generator and discriminator progressively by adding new layers to both the generator and discriminator as the training progresses. In this work, we propose a progressively unfreezing scheme for PUPGAN, which is quite different from the previous approaches.

Similar to GANs, PUPGAN can be considered as a general framework for generating images with fine texture details. It can be used for various image generation tasks. Among others, in this work, we have applied PUPGAN to the unpaired image-to-image translation task, which is a very challenging image generation task. Recently, CycleGAN has been applied to this task~\cite{DBLP:conf/iccv/ZhuPIE17}. However, the generated images are generally lack of texture details. On the contrary, with the adaptive perceptual discriminator, PUPGAN can perform well on the unpaired image-to-image translation task.
\section{Progressively Unfreezing Perceptual GAN}
In this section, we introduce the proposed Progressively Unfreezing Perceptual GAN (PUPGAN) in detail. We first present an overview of the PUPGAN framework. Then, we describe the architecture of the adaptive perceptual discriminator in PUPGAN. Next, we introduce the progressively unfreezing scheme for the training of PUPGAN. Finally, we specify how to apply PUPGAN to the unpaired image-to-image translation task.
\subsection{The PUPGAN Framework}
PUPGAN consists of a generator $G$ and an adaptive perceptual discriminator $D^{ap}$. The generator $G$ is trained to learn a mapping from the input distribution $x\sim p_x$ to the real data distribution $p_{data}$. For the architecture of the generator $G$, we just take the generator of the baseline method. Namely, there are no modifications to the generator $G$. The adaptive perceptual discriminator $D^{ap}$, which consists of a perceptual feature extractor and the discriminative learning layers, is trained to measure the multi-level feature discrepancy between the real images $y \sim p_{data}$ and the generated images $G(x)$. The learning objective can be written as follows:

\begin{eqnarray}\label{eq:gan}
\begin{aligned}
\min \limits_{G}\max \limits_{D^{ap}}{V(D^{ap},G)}=&E_{y\sim p_{data}}[\log D^{ap}(y)]\\
&+E_{x\sim p_x}[\log(1-D^{ap}(G(x)))].
\end{aligned}
\end{eqnarray}
The generator $G$ and the adaptive perceptual discriminator $D^{ap}$ are trained adversarially to compete with each other. The generator is encouraged to simulate the real images $y$ fed with the input images $x$. The adaptive perceptual discriminator is trained to distinguish the generated images $\tilde{y}$~($G(x)$) from the real images $y$. Hence, the adversarial loss can be defined as:
\begin{eqnarray}\small\label{eq:rg}
L_{Adv}=-E_{x\sim p_G(x)}[\log(D^{ap}(G(x)))].
\end{eqnarray}
The adversarial loss encourages the generated distributions to reside on the manifold of the real images by penalizing the discrepancy between the generated and real images.

\subsection{The Adaptive Perceptual Discriminator}
Unlike previous GANs using an external classifier to compute the perceptual loss, we use the discriminator of PUPGAN to evaluate the feature discrepancy between the generated and real images.

\begin{figure}
  \centering
  \includegraphics[scale=0.14]{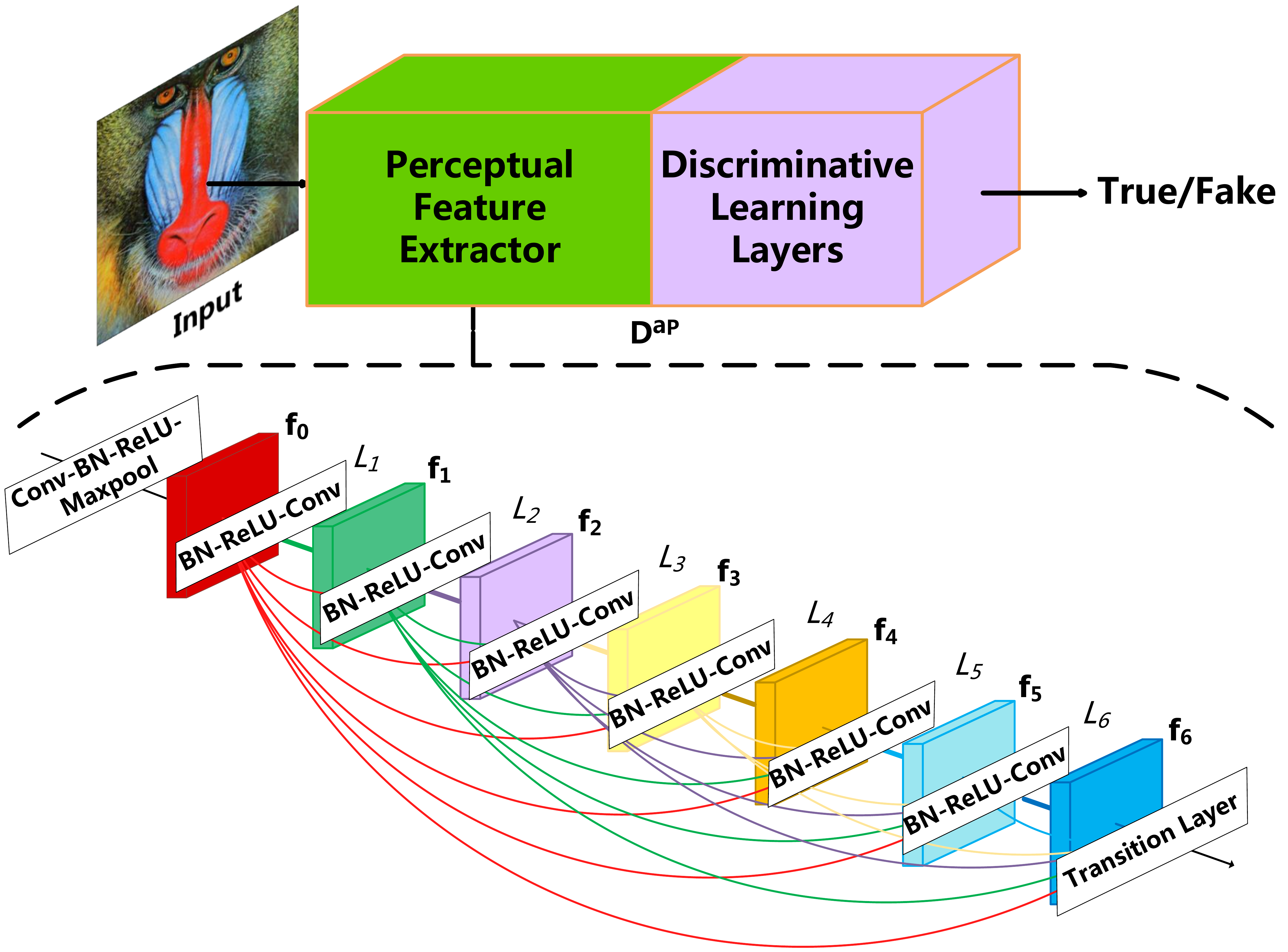}
  \caption{The adaptive perceptual discriminator of PUPGAN.}
  \label{denseblock}
\end{figure}
In this paper, we propose an adaptive perceptual discriminator, which is composed of a perceptual feature extractor and some discriminative learning layers (as shown in Fig. \ref{denseblock}). Specifically, the perceptual feature extractor employs the hidden layer of the discriminator to extract the perceptual features and the discriminative learning layers are used to measure the feature discrepancy. More importantly, the perceputal feature extractor in the adaptive perceptual discriminator can be learned along with the training of the PUPGAN.

The perceptual feature extractor consists of a $7 \times 7$ convolution, a $3 \times 3$ max pooling, and a 6-layer dense block where each layer consists of a sequence of $1 \times 1$ convolutions and a $3 \times 3$ convolutions. Before each convolution, a batch normalization and the ReLU activation are adopted. Specifically, the 6-layer dense block is taken from Dense-121~\cite{DBLP:conf/cvpr/HuangLMW17}, which is pre-trained on the ImageNet classification task~\cite{DBLP:journals/ijcv/RussakovskyDSKS15}. The dense skip connections guarantee the forwarding of multi-level features to the discriminative learning layers. Furthermore, for error back propagation, the skip connections ensure the gradients from the discriminative learning layers and unfreezed layers to pass on to the generator. Hence, the discriminative learning layers measure a multi-level discrepancy between the generated and real images. 

The perceptual feature extractor in the discriminator performs a transfer learning for the current image generation task. As a result, the perceptual features extracted by the adaptive perceptual discriminator is more suitable for the current task than the traditional perceptual loss and models dirctly embeded the VGGNet into the discriminators. In addition, compared to the traditional perceptual loss with element-wise measuring methods, using the discriminative learning layers can better measure the discrepancy of the perceptual features.

\subsection{Progressively Unfreezing Scheme}\label{sec:pud}
\subsubsection{Three Ways for Employing the Feature Extractor}
The pre-trained perceptual feature extractor in the adaptive perceptual discriminator can be employed in three ways. The first way is to update all the parameters in the perceptual feature extractor with the entire model, which we called normal transformation. The second way is to freeze all the parameters in the perceptual feature extractor, which we called no transformation (same as \cite{DBLP:conf/eccv/SungatullinaZUL18}). The third way is a compromise between them, which progressively unfreezes the parameters in the perceptual feature extractor layer-by-layer. We call it progressive transformation.

For the first way, at the beginning of training, since the discriminator has strong discriminative capability, it is trial to find the split hyperplane between the generated images and the real images, whereas the generator with low generative capability cannot fool the discriminator (as mentioned in Section \ref{sec:intro}). As a result, the gradient produced by the discriminator may lead to the gradient vanishing or mode collapse problems.

For the second way, when the parameters of the pre-trained perceptual feature extractor are freezed, its learning capability is restricted. In other words, the provided perceptual feature extractor is pre-trained on the ImageNet classification task, its performance is limited on the current image perceptual feature extraction task.

Here, we choose the third way, i.e. progressive transformation, which progressively unfreezes the pre-trained perceptual feature extractor in the adaptive perceptual discriminator.
\vspace{-0.2cm}
\subsubsection{Progressively Unfreezing with the Unfreezing Factor $\varphi$}
To achieve the progressive transformation, we propose a progressively unfreezing scheme, which depends on an unfreezing factor $\varphi$. 

The training process is shown in Algorithm \ref{alg:1}. At the beginning of training, the parameters of the perceptual extractor are all freezed. As the training advances, the parameters of the bottom layer in the perceptual feature extractor that connects to the discriminative part are first unfreezed. Next, we progressively unfreeze the front layers one-by-one depending on a randomly generated probability. If the probability larger than a threshold (we empirically set the threshold $\varphi = 0.66$, please refer to the supplementary material), we unfreeze the parameters of one convolutional layer in front of the unfreezed layer. Namely, the unfreezed layers in the perceptual feature extractor and the discriminative learning layers remain trainable, while the parameters of the other layers in the perceptual feature extractor are freezed. Hence, the perceputal feature extractor smoothly adapt to the current image generation task.
\begin{algorithm}
\caption{The progressively unfreezing scheme.}
\label{alg:1}
 \begin{algorithmic}[1]
  \REQUIRE $\theta_G$: $G$'s parameters; $\theta_p$: perceptual feature extractor's parameters: $\theta_d$: discriminative learning layers' parameters; $m$: batchsize; $\varphi$: unfreezing factor (empirically set to 0.66).
  \REQUIRE Adam hyperparameters: $\alpha_p=1\times10^{-6}$, $\alpha_d=2\times10^{-4}$, $\beta_1=0.5$ and $\beta_1=0.999$.
  \STATE Freeze all of the parameters $\theta_p$
  \FOR {e=1, 2, ..., epoch}
       \STATE $p=$ random($1$)
       \IF {$p>\varphi$}
           \STATE Unfreeze one layer in the perceptual feature extractor which is the nearest neighbor to the discriminative learning layers.
       \ENDIF
       \FOR {j=1, 2, ..., datasize$/m$}
            \STATE Sample {$y_1$, \ldots, $y_m$} from the real data distribution
            \STATE Sample {$x_1$, \ldots, $x_m$} from the input data distribution

            \STATE $L_{D^{ap}}\leftarrow \frac {1}{m} \sum[log D^{ap}(y_i)+log(1-D^{ap}(G(x_i)))]$
            \STATE $L_G\leftarrow \frac {1}{m} \sum[log D^{ap}(G(x_i))]$
            \STATE $\theta_d, \theta_p = Adam(L_{D^{ap}}, \theta_d, \theta_p, \alpha_d, \alpha_p, \beta_1, \beta_2)$
            \STATE $\theta_G = Adam(L_{D^{ap}}, \theta_G, \beta_1, \beta_2)$
       \ENDFOR
  \ENDFOR
 \end{algorithmic}
\end{algorithm}

There are several benefits for our progressively unfreezing scheme. First, the generation at the beginning is stable, because the adaptive perceptual discriminator with most parameters freezed is easy to fool by the generator. As the training progresses, the generator is asked for a more complicated question than the previous one by unfreezing the parameters in the adaptive perceptual discriminator. In addition, it is a reliable progressively training scheme, which can successfully avoid the gradient vanishing and model collapse problems.

Another benefit is that unfreezing the parameters of the perceptual feature extractor layer-by-layer can reduce the training time. Unlike normal transformation, when progressively unfreeze the parameters of the perceptual feature extractor, most parameters of the convolutional layers are freezed without update. In other words, the pre-trained feature extractor is incrementally transformed to that for the current task. In contrast to normal transformation, which needs to update all the parameters in the perceptual feature extractor, the progressive transformation method is more efficient.

In addition, the dense block is extremely suitable to the progressively unfreezing scheme. The skip connections between layers make it possible for information flow between the layers. Moreover, compared with VGGNet, the dense block has fewer parameters and better feature extraction capability.

\subsection{PUPGAN for Unpaired Image-to-Image Translation}
In GANs' research field, paired examples are generally required to train the network, as the discriminator needs both the generated images and the real images. CycleGAN \cite{DBLP:conf/iccv/ZhuPIE17} made a breakthrough on the unpaired image-to-image translation task, which introduced a cycle consistency loss to measure the discrepancy between the inputs with the reversely generated images. Although it has achieved competitive results on image-to-image translation task, the generated images are usually lack of texture details. Besides, the traditional perceptual loss, which needs both real images and generated images to measure the feature discrepancy, cannot work on the tasks without the pair-wise images.

The adaptive perceptual discriminator in PUPGAN extracts the multi-level perceptual features and measures the discrepancy between them. It overcomes the shortcomings of the traditional perceptual loss that needs pair-wise images to measure the feature discrepancy. Furthermore, the adaptive perceptual discriminator in PUPGAN efficiently extract and measure the multi-level feature discrepancy between the generated images and the real images. Hence, it can lead to better texture details and perceptual characteristics than the images generated by CycleGAN.

\section{Experiments}
To verify the universal validity of PUPGAN, we evaluate the performance of PUPGAN on several image generation tasks. Specifically, we choose three representative image generation tasks, i.e. single image super-resolution, paired image-to-image translation and unpaired image-to-image translation, for each task, we choose the classical baseline methods for comparison. For any other state-of-the-art GANs-based model, we can replace its discriminator to the adaptive perceptual discriminator designed in this work, and easily compare with it on the learning tasks.

In this section, we show some representative experimental results. For more implementation details and experimental results, please refer to the supplementary materials.

\subsection{Datasets and Evaluation Metrics}
For the single image super-resolution, we evaluated PUPGAN on several widely used benchmark datasets, i. e. Set5~\cite{DBLP:conf/bmvc/BevilacquaRGA12}, Set14~\cite{DBLP:conf/cas/ZeydeEP10}, BSD100~\cite{DBLP:conf/iccv/MartinFTM01} and Urban100~\cite{DBLP:conf/cvpr/HuangSA15}. All experiments were performed with a scale factor of $4 \times$ between the low-resolution and high-resolution images. Meanwhile, for quantitative comparison, we evaluated the generated images in terms of peak signal to noise ratio (PSNR) and structural similarity idex (SSIM) \cite{DBLP:journals/tip/WangBSS04}.

For the paired image-to-image translation, we conducted experiments on several tasks, such as aerial$\rightarrow$map with training data scraped from Google Maps~\cite{DBLP:conf/cvpr/IsolaZZE17}, label$\rightarrow$facade with images from the CMP facade dataset \cite{DBLP:conf/dagm/TylecekS13}. Moreover, we evaluated the generated images by PUPGAN and the compared methods in terms of PSNR and SSIM.

For the unpaired image-to-image translation, we chose several tasks performed in CycleGAN \cite{DBLP:conf/iccv/ZhuPIE17}. The tasks included aerial$\leftrightarrow$map~\cite{DBLP:conf/cvpr/IsolaZZE17}, label$\leftrightarrow$facade~\cite{DBLP:conf/dagm/TylecekS13}, horse$\leftrightarrow$zebra and orange$\leftrightarrow$apple with images downloaded from ImageNet \cite{DBLP:journals/ijcv/RussakovskyDSKS15} and summer$\leftrightarrow$winter Yosemite photos from Flickr (\url{http://www.flickr.com}). In addition, for quantitative comparison, we use the perceptual quality index (PQI)~\cite{DBLP:conf/cvpr/BlauM18} to measure the quality of the generated images. 

\subsection{Parameter Settings}
To demonstrate the effectiveness of PUPGAN, we conducted the ablation study. The adaptive perceptual discriminator in PUPGAN is represented as ``Dense\_D". For the perceptual feature extractor, we took a pre-trained VGGNet-19 as the variant, represented as ``VGG\_D". For the training scheme, we considered to progressively unfreeze the parameters in the perceptual extractor layer-by-layer or freeze all the parameters of the perceptual feature extractor, represented as ``UF" or ``F" respectively. While unfreezing all the parameters at the beginning of training may cause the gradient vanishing or mode collapse problems, it is not considered here. In addition, for the regularization mechanism, we conducted experiments that with or without spectral normalization \cite{DBLP:conf/iclr/MiyatoKKY18} after each convolutional layer in the discriminator. The spectral normalization is represented as ``SN".

The generators of PUPGAN for the applied tasks are the same as that of the baseline models. The adaptive perceptual discriminators of PUPGAN for different tasks are designed with their unique discriminative learning layers. The specific architectures are described in the following subsections and in the supplementray material.

\subsection{Single Image Super-Resolution}
For the single image super-resolution task, we take SRGAN~\cite{DBLP:conf/cvpr/LedigTHCCAATTWS17} as the baseline. SRGAN is a famous GANs-based model for single image super-resolution. It utilizes a VGGNet loss which is based on a pre-trained VGGNet-19 network, and measures the high-level discrepancy with the Euclidean distance.

\begin{figure}[h]
  \centering
  \includegraphics[scale=0.75]{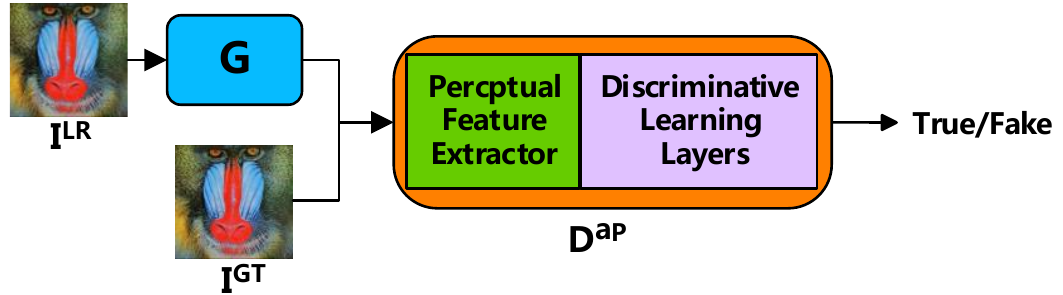}
  \caption{The architecture of PUPGAN for single image super-resolution.}
  \label{fig:sr}
\end{figure}
Fig. \ref{fig:sr} illustrates the architecture of PUPGAN for single image super-resolution. The discriminative learning layers consist of three $3\times3$ convolutions with stride $2$ followed by average pooling and two $1\times1$ convolutions with stride $1$. The convolutional layers except the last one are followed by a LeakyReLU activation ($\alpha = 0.2$) and spectral normalization.

\begin{figure}[t]
  \centering
  \includegraphics[width=\textwidth]{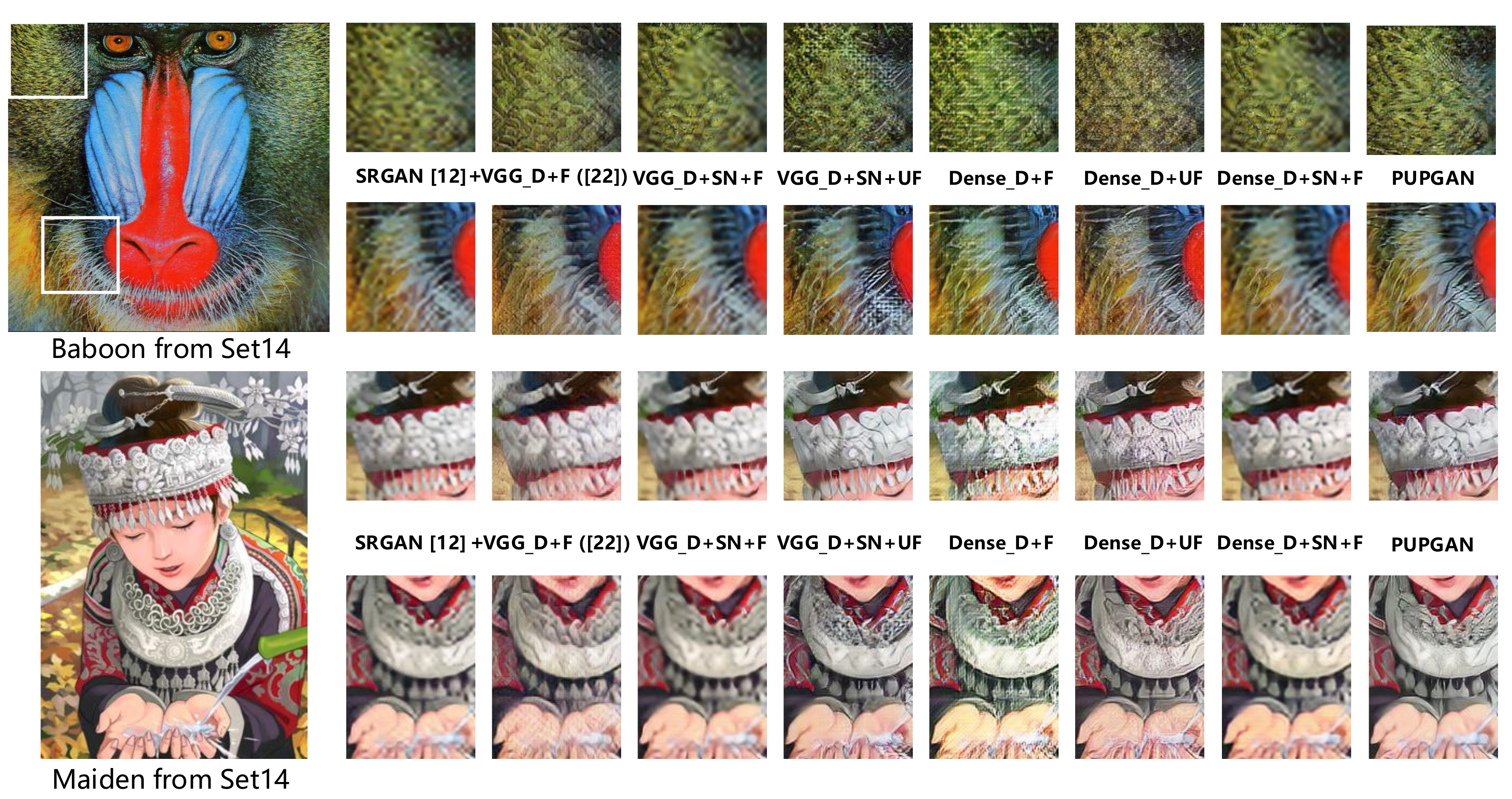}
  \caption{From left to right: ground-truth, SRGAN~\cite{DBLP:conf/cvpr/LedigTHCCAATTWS17}, SRGAN + VGGNet-based discriminator with freezed parameters, SRGAN + VGGNet-based discriminator + SN with freezed parameters, SRGAN + VGGNet-based discriminator + SN with progressively unfreezing scheme, SRGAN + adaptive perceptual discriminator with freezed parameters, SRGAN + adaptive perceptual discriminator with unfreezed parameters, SRGAN + adaptive perceptual discriminator + SN with freezed parameters and PUPGAN.}
  \label{srgan}
\end{figure}

Fig. \ref{srgan} demonstrates the $4\times$ upscaling super-resolved images generated by the baseline SRGAN, PUPGAN and the other compared approaches. It is obvious that the images generated by SRGAN are blurry and without enough texture details. In general, the images generated by the models with ``VGG\_D" is more blurry than those with ``Dense\_D". Furthermore, utilizing progressively unfreezing scheme further enhanced the generative capability of the networks. It is worth mentioning that PUPGAN achieved the best performance. It not only recovered the texture details of the images, but also kept the textures approximate to the real texture in the ground-truth images.

\begin{table}[t]
\centering
\setlength{\abovecaptionskip}{0 pt}
\caption{Performance of the baseline SRGAN~\cite{DBLP:conf/cvpr/LedigTHCCAATTWS17}, PUPGAN and other variants for single image super-resolution in terms of PSNR/SSIM on the Set5, Set14, BSD100 and Uber100 datasets.}
\label{SRGAN}
\scalebox{0.8}[0.8]{
\begin{tabular}{c|cc|cc|cc|cc}
  \hline
  \thead{ \backslashbox{Methods}{Datasets}}
  &\multicolumn{2}{c}\centering\thead{Set5~\cite{DBLP:conf/bmvc/BevilacquaRGA12}}&\multicolumn{2}{c}\centering\thead{Set14~\cite{DBLP:conf/cas/ZeydeEP10}}
  &\multicolumn{2}{c}\centering\thead{BSD100~\cite{DBLP:conf/iccv/MartinFTM01}}&\multicolumn{2}{c}\centering\thead{Uber100~\cite{DBLP:conf/cvpr/HuangSA15}}\\
  \hline
  \thead{   }&\thead{ PSNR  }&\thead{SSIM}&\thead{ PSNR  }&\thead{SSIM}&\thead{ PSNR  }&\thead{SSIM}&\thead{ PSNR  }&\thead{SSIM}\\
  \thead{SRGAN~\cite{DBLP:conf/cvpr/LedigTHCCAATTWS17} }&\thead{ 28.5694 }&\thead{ 0.8271 }&\thead{25.4943  }&\thead{ 0.7266 }&\thead{ 25.6282 }&\thead{0.6935 }&\thead{ 23.4113 }&\thead{ 0.7099 }\\

  \thead{SRGAN~\cite{DBLP:conf/cvpr/LedigTHCCAATTWS17}+VGG\_D+F (\cite{DBLP:conf/eccv/SungatullinaZUL18})} &\thead{ 24.2069 } &\thead{0.6185  }&\thead{ 22.7773 } &\thead{ 0.5455 }&\thead{22.8579 } &\thead{ 0.5262 }&\thead{ 21.3465} &\thead{0.5761 }\\
  \thead{SRGAN~\cite{DBLP:conf/cvpr/LedigTHCCAATTWS17}+VGG\_D+SN+F} &\thead{ 27.5899 } &\thead{ 0.7686 }&\thead{25.1665 } &\thead{0.7249  }&\thead{ 25.3625} &\thead{ 0.6769 }&\thead{23.0150  } &\thead{ 0.7020 }\\
  \thead{SRGAN~\cite{DBLP:conf/cvpr/LedigTHCCAATTWS17}+VGG\_D+SN+UF } &\thead{ 24.5858 } &\thead{ 0.7386 }&\thead{22.1574  } &\thead{ 0.6247 }&\thead{ 22.8857 } &\thead{ 0.6134 }&\thead{ 20.5188 } &\thead{ 0.6169 }\\

  \thead{SRGAN~\cite{DBLP:conf/cvpr/LedigTHCCAATTWS17}+Dense\_D+F} &\thead{ 23.3219 } &\thead{ 0.6699 }&\thead{ 21.4114 } &\thead{0.5920  }&\thead{  21.8136} &\thead{ 0.5455 }&\thead{19.7500  } &\thead{ 0.5463 }\\
  \thead{SRGAN~\cite{DBLP:conf/cvpr/LedigTHCCAATTWS17}+Dense\_D+SN+F} &\thead{ 27.6365 } &\thead{0.8092  }&\thead{25.0694  } &\thead{0.7184 }&\thead{ 25.1767 } &\thead{ 0.6832 }&\thead{22.9457 } &\thead{ 0.6892 }\\
  \thead{SRGAN~\cite{DBLP:conf/cvpr/LedigTHCCAATTWS17}+Dense\_D+UF } &\thead{ 24.3997 } &\thead{ 0.6643 }&\thead{22.4456  } &\thead{ 0.5933 }&\thead{ 22.5424 } &\thead{ 0.5569 }&\thead{ 20.9289 } &\thead{ 0.6037 }\\
  \hline
  \thead{PUPGAN} &\thead{ $\boldsymbol{29.0598}$ } &\thead{ $\boldsymbol{0.8495}$ }&\thead{$\boldsymbol{25.7471}$  } &\thead{ $\boldsymbol{0.7454}$ }&\thead{$\boldsymbol{25.8316}$  } &\thead{ $\boldsymbol{0.7074}$ }&\thead{ $\boldsymbol{23.5533}$ } &\thead{ $\boldsymbol{0.7297}$ }\\
  \hline
\end{tabular}}
\end{table}
Table \ref{SRGAN} illustrates the quantitative results on several benchmark datasets. PUPGAN obtained the highest score among the other compared approaches. Even though SRGAN received the second-highest score, the images generated by it is blurry (as shown in Fig. \ref{srgan}). 

\subsection{Paired Image-to-Image Translation}\label{sec:pix}
For the paired image-to-image task, we take Pix2Pix~\cite{DBLP:conf/cvpr/IsolaZZE17} as the baseline. Pix2Pix is a generic tool for paired image-to-image translation tasks, which utilized a conditional GAN-based model to achieve the reasonable results. It optimizes the entire model with a generative adversarial loss and an MAE loss between the generated images and the real images. 

\begin{figure}[h]
  \centering
  \includegraphics[scale=0.7]{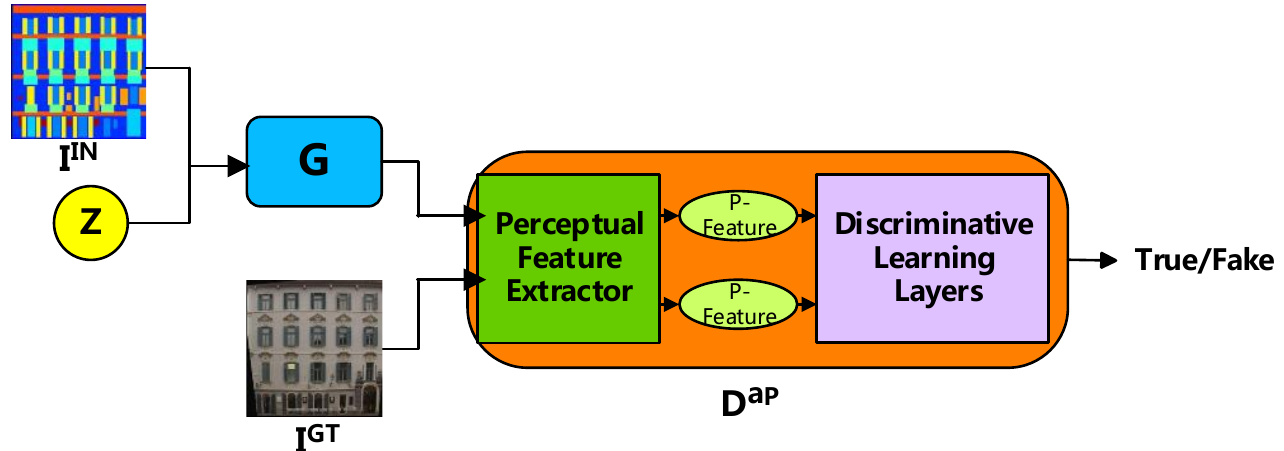}
  \caption{The architecture of PUPGAN for paired image-to-image translation.}
  \label{fig:pix}
\end{figure}
Fig. \ref{fig:pix} illustrates the structure of PUPGAN for paired image-to-image translation. Notice that in the Pix2Pix model, the generated image and the real image are concatenated as the input to the discriminator. While for the adaptive perceptual discriminator, the generated image and the real image first go through the perceptual feature extractor, and then the perceptual features are concatenated as the input to the discriminative learning layers. Here, the discriminative learning layers consist of four $3\times3$ convolutions with stride $2$ and a $4\times4$ convolution with stride $1$. The convolutional layers are followed by a LeakyReLU activation ($\alpha = 0.2$) and spectral normalization.

\begin{figure}[t]
  \centering
  \includegraphics[width=\textwidth]{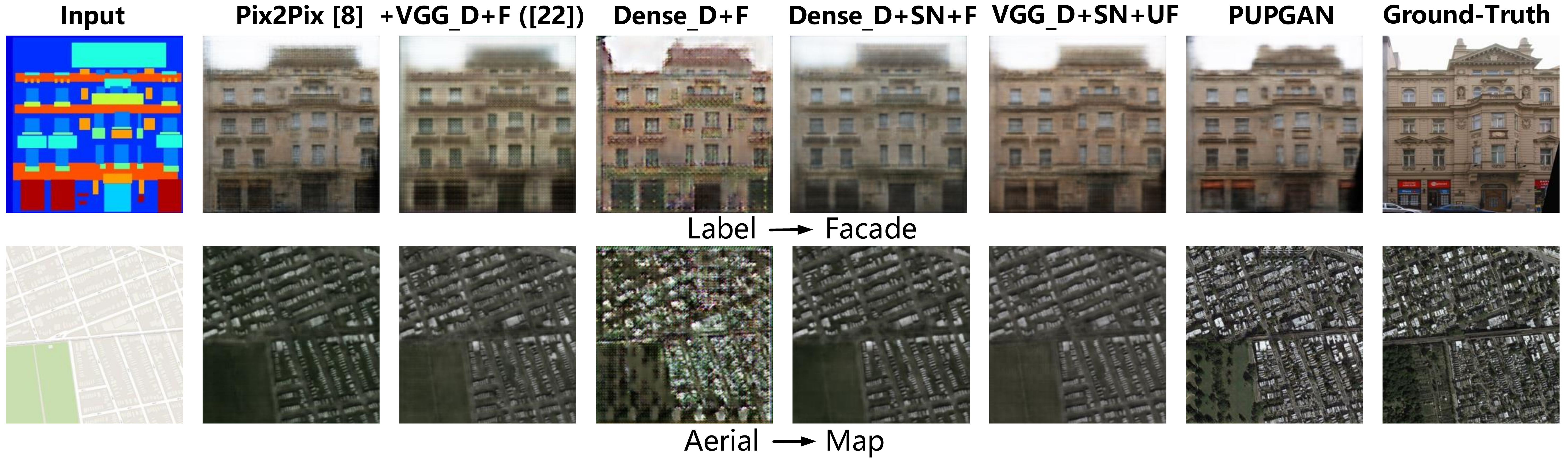}
  \caption{The comparison between different methods for the paired image translation task. Pix2Pix~\cite{DBLP:conf/cvpr/IsolaZZE17} is the baseline method.}
  \label{pixf}
\end{figure}

Fig. \ref{pixf} shows the images generated by the baseline Pix2Pix, PUPGAN and the other compared approaches on the label$\rightarrow$facade and aerial$\rightarrow$map tasks. The image details generated by PUPGAN is much better than the other compared approaches. It should be noted that the images generated by PUPGAN are the closest to the ground-truth image, no matter the color of the objects or the fine texture details.

\begin{table}[h]
\centering
\setlength{\abovecaptionskip}{0pt}
\caption{Performance of PUPGAN, the baseline Pix2Pix~\cite{DBLP:conf/cvpr/IsolaZZE17} and other variants for the paired image translation tasks in terms of PSNR and SSIM.}
\label{pix:tab}
\scalebox{0.8}[0.8]{
\begin{tabular}{c|cc|cc}
  \hline
  \thead{ \backslashbox{Methods}{Datasets}}
  &\multicolumn{2}{c}\centering\thead{Label$\rightarrow$Facade}&\multicolumn{2}{c}\centering\thead{Aerial$\rightarrow$Map}\\
  \hline
  \thead{   }&\thead{ PSNR  }&\thead{SSIM}&\thead{ PSNR  }&\thead{SSIM}\\
  \thead{Pix2Pix~\cite{DBLP:conf/cvpr/IsolaZZE17}}&\thead{ 18.8356 }&\thead{ 0.4955}&\thead{ 17.4289 }&\thead{ 0.3749 }\\
  \thead{Pix2Pix~\cite{DBLP:conf/cvpr/IsolaZZE17}+VGG\_D+F (\cite{DBLP:conf/eccv/SungatullinaZUL18})} &\thead{17.4117}&\thead{0.3691}&\thead{18.2702}&\thead{0.3954}\\
  \thead{Pix2Pix~\cite{DBLP:conf/cvpr/IsolaZZE17}+Dense\_D+F} &\thead{16.8993}&\thead{0.3560}&\thead{15.2598}&\thead{0.3554}\\
  \thead{Pix2Pix~\cite{DBLP:conf/cvpr/IsolaZZE17}+VGG\_D+SN+F}  &\thead{18.8696}&\thead{0.4794}&\thead{18.2852}&\thead{0.3971}\\
  \thead{Pix2Pix~\cite{DBLP:conf/cvpr/IsolaZZE17}+Dense\_D+SN+F}  &\thead{19.4678}&\thead{0.5234}&\thead{17.8319}&\thead{0.3980}\\ 
  \thead{Pix2Pix~\cite{DBLP:conf/cvpr/IsolaZZE17}+VGG\_D+SN+UF }  &\thead{18.6036}&\thead{0.4902}&\thead{17.3374}&\thead{0.3668}\\
  \hline
  \thead{PUPGAN} &\thead{ $\boldsymbol{21.3647}$ } &\thead{ $\boldsymbol{0.6335}$ }&\thead{$\boldsymbol{19.7952}$  } &\thead{ $\boldsymbol{0.48649}$ }\\
  \hline
\end{tabular}}
\end{table}
Table \ref{pix:tab} illustrates the quantitative results for the paired image translation tasks. PUPGAN obtained the highest score among the other compared approaches. In addition, we can see that taking the dense block as the perceptual feature extractor performs better than that taking the VGGNet as the perceptual feature extractor (from the last two lines in Table \ref{pix:tab}). It demonstrates that the dense connections in the feature extractor is extremely suitable to the progressively unfreezing scheme. 

\subsection{Unpaired Image-to-Image Translation}
For the unpaired image-to-image translation task, we take CycleGAN~\cite{DBLP:conf/iccv/ZhuPIE17} as the baseline. CycleGAN is a well-known tool for unpaired image-to-image translation task. It contains both forward and reverse mapping functions between the source domain and target domain. It is optimized with two adversarial losses and a cycle consistency loss. However, in this case, the traditional perceptual loss cannot work, as the target image corresponding to a generated image is unknown. Although Sungatullina et al.~\cite{DBLP:conf/eccv/SungatullinaZUL18} embeded a pre-trained VGGNet inside the discriminator to extract the high-level features, the parameters in the pre-trained network were fixed, whereas the adaptive perceptual discriminator can help to learn the suitable high-level perceptual features adapted to the current task and generate images with fine texture details. 

\begin{figure}[h]
  \centering
  \includegraphics[scale=0.7]{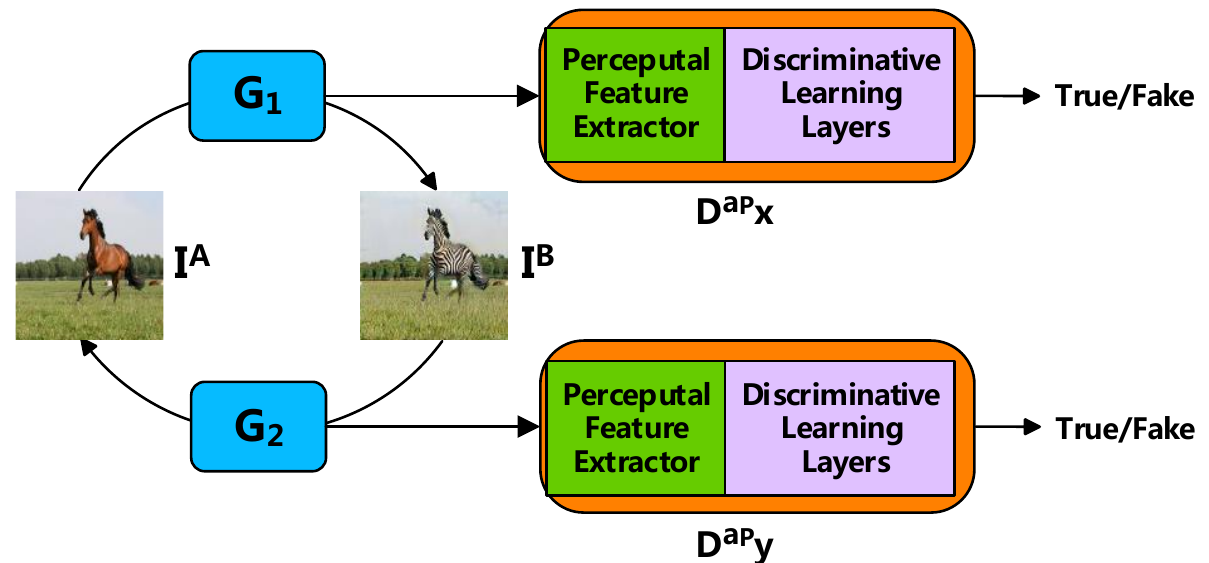}
  \caption{The architecture of PUPGAN for unpaired image-to-image translation.}
  \label{fig:cycle}
\end{figure}
Fig. \ref{fig:cycle} illustrates the structure of PUPGAN for unpaired image-to-image translation. The structures of the discriminative learning layers are as same as we illustrated in Section \ref{sec:pix}.

\begin{figure}
  \centering
  \includegraphics[width=\textwidth]{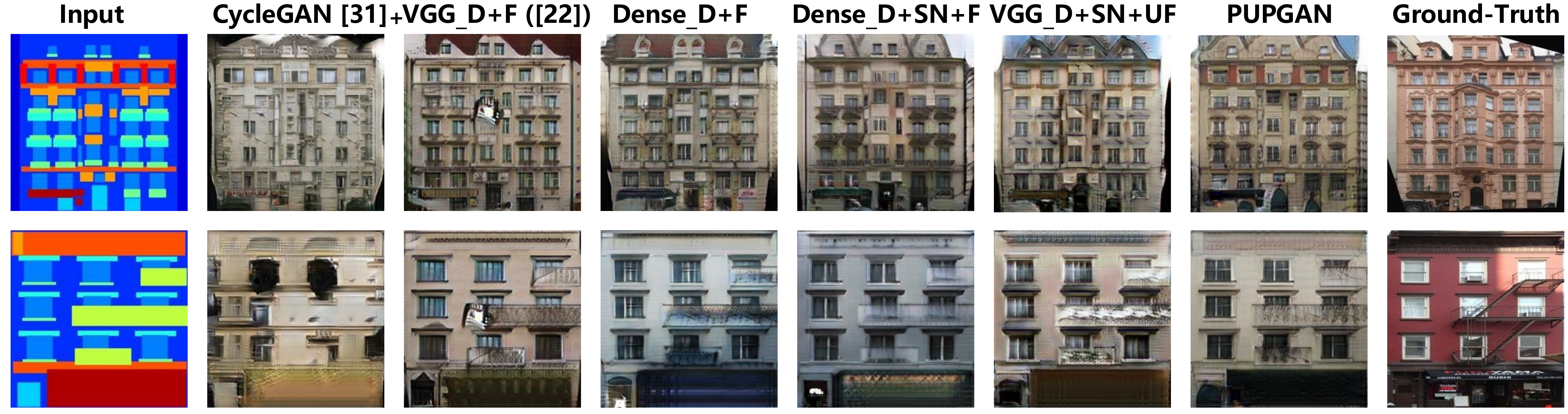}
  \caption{The comparison for the unpaired image-to-image translation on label$\rightarrow$facade task. We took CycleGAN~\cite{DBLP:conf/iccv/ZhuPIE17} as the baseline and conduct extensive experiments on the variants of PUPGAN.}
  \label{cyclef}
\end{figure}

We first test the performance of PUPGAN and the compared approaches for unpaired image-to-image translation on paired datasets, where ground-truth input-output pairs are available for evaluation. For training, we shuffled the images to made them unpaired. Fig. \ref{cyclef} shows the label$\rightarrow$facade results obtained by PUPGAN and the other compared approaches. It is easy to see that, the images generated by PUPGAN are much better than that generated by CycleGAN and the compared approaches, whether from the contour of the facades or the texture details in the generated images.

\begin{figure}
  \centering
  \includegraphics[width=\textwidth]{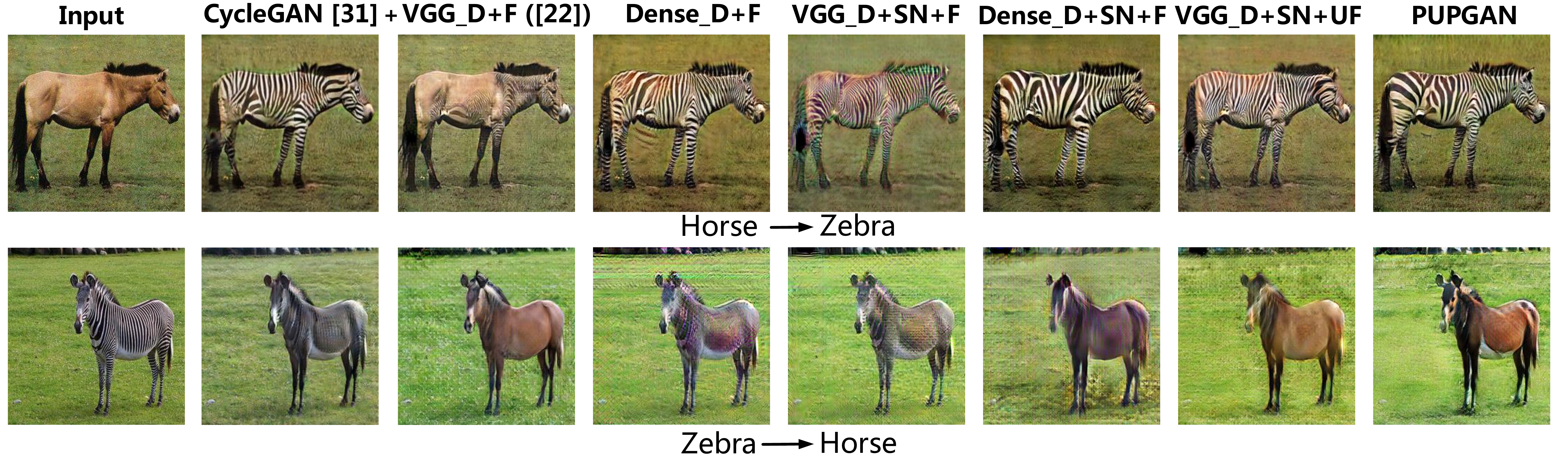}
  \caption{The comparison for the unpaired image-to-image translation on horse$\leftrightarrow$zebra task, where paired data does not exist.}
  \label{cycleh}
\end{figure}
In addition, we conducted experiments on the unpaired image-to-image translation task, horse$\leftrightarrow$zebra. As shown in Fig. \ref{cycleh}, images generated by PUPGAN have fine texture details and correct color. Specifically, even the zebra's mane was painted in white and black by PUPGAN, while the compared approaches did not have this property.

\begin{table}
\centering
\setlength{\abovecaptionskip}{0pt}
\caption{Performance of PUPGAN, the baseline CycleGAN~\cite{DBLP:conf/iccv/ZhuPIE17} and other variants for unpaired image translation tasks in terms of PQI.}
\label{cycle:tab}
\scalebox{0.68}[0.68]{
\begin{tabular}{c|c|c|c|c|c|c|c}
\hline
\thead{PQI}&\thead{CycleGAN~\cite{DBLP:conf/iccv/ZhuPIE17}} & \thead{VGG\_D+F~(\cite{DBLP:conf/eccv/SungatullinaZUL18})} & \thead{Dense\_D+F}& \thead{VGG\_D+SN+F} & \thead{VGG\_D+SN+UF}  & \thead{Dense\_D+SN+F} & \thead{PUPGAN}\\
\hline
\thead{Label$\rightarrow$Facade} &\thead{5.0085}&\thead{4.8144}&\thead{5.8095}&\thead{4.4230}&\thead{4.1031}&\thead{5.2549}& \thead{\bf4.0308}\\
\hline
\thead{Facade$\rightarrow$Label} &\thead{6.6129}&\thead{4.9482}&\thead{5.2549}&\thead{4.6094}&\thead{4.4423}&\thead{4.4885}& \thead{\bf4.3692} \\
\hline
\thead{Horse$\rightarrow$Zebra} &\thead{3.3219}&\thead{3.2865}&\thead{3.2363}&\thead{3.4095}&\thead{3.5321}&\thead{3.2862}&\thead{\bf2.8842}\\
\hline
\thead{Zebra$\rightarrow$Horse} &\thead{3.3278}&\thead{3.2679}&\thead{3.2979}&\thead{3.525}&\thead{3.2938}&\thead{3.2862}&\thead{\bf3.2109}\\
\hline
\end{tabular}}
\end{table}
Table \ref{cycle:tab} shows the quantitative results for the label$\leftrightarrow$facade and horse$\rightarrow$zebra tasks in terms of PQI. The lower the PQI value, the higher the perceptual quality recovered by the method. It obvious that PUPGAN obtained the best result among the other compared methods. For more experimental results, please refer to the supplementary materials.


\section{Conclusion}
In this paper, we propose a new framework called PUPGAN, which can generate images with fine texture details. Its adaptive perceptual discriminator, which utilizes a pre-trained dense block as the perceptual feature extractor and transfers it to the current task, efficiently measures the multi-level discrepancy between the generated and real images. In addition, we propose a progressively unfreezing scheme to smoothly improve the generator's image generation capability. The qualitative and quantitative experiments demonstrate the effectiveness of PUPGAN for texture details generation on three representative image generation tasks, i.e. single image super-resolution, paired image-to-image translation and unpaired image-to-image translation. Last but not the least, PUPGAN can be considered as a general framework for image generation with fine texture details, it can be applied to more image generation tasks other than that we performed here.
%
%
\bibliographystyle{splncs04}
\bibliography{egbib}
\end{document}